\documentclass[10pt,journal,compsoc]{IEEEtran}
\usepackage{graphicx}
\usepackage{subfigure}

\usepackage{amsmath}
\usepackage{float}
\usepackage{color}

\usepackage{xspace}
\usepackage{multirow}
\usepackage{booktabs}
\usepackage[misc]{ifsym}
\usepackage{amssymb}

\usepackage{subfigure}
\usepackage{booktabs}

\usepackage{paralist, tabularx}
\usepackage{multirow} 

\usepackage[colorlinks,linkcolor=blue]{hyperref}

\ifCLASSOPTIONcompsoc
  \usepackage[nocompress]{cite}
\else
  \usepackage{cite}
\fi
\ifCLASSINFOpdf
\else
\fi

\begin{document}

\title{Enhanced Spatio-Temporal Interaction Learning for Video Deraining: Faster and Better}

\author{
Kaihao Zhang, Dongxu Li, \ Wenhan Luo,  \ Wenqi Ren,
\ and Wei Liu \\
\IEEEcompsocitemizethanks{\IEEEcompsocthanksitem K. Zhang and D Li are with the College of Engineering and Computer Science, the Australian National University, Canberra, Australia.\protect\\
E-mail: \{kaihao.zhang, dongxu.li\}@anu.edu.au
\IEEEcompsocthanksitem W. Luo and W. Liu are with the Tencent, Shenzhen 518057, China.
E-mail:\{whluo.china@gmail.com; wl2223@columbia.edu.\protect \}
\IEEEcompsocthanksitem W. Ren is with State Key Laboratory of Information Security, Institute of Information Engineering, Chinese Academy of Sciences, Beijing, 100093, China. E-mail: rwq.renwenqi@gmail.com. \protect
\IEEEcompsocthanksitem Corresponding author: Dongxu Li
}}

\IEEEtitleabstractindextext{
\begin{abstract}
Video deraining is an important task in computer vision as the unwanted rain hampers the visibility of videos and deteriorates the robustness of most outdoor vision systems. Despite the significant success which has been achieved for video deraining recently, two major challenges remain: 1) how to exploit the vast information among successive frames to extract powerful spatio-temporal features across both the spatial and temporal domains, and 2) how to restore high-quality derained videos with a high-speed approach. 
In this paper, we present a new end-to-end video deraining framework, dubbed Enhanced Spatio-Temporal Interaction Network (ESTINet), which considerably boosts current state-of-the-art video deraining quality and speed. The ESTINet takes the advantage of deep residual networks and convolutional long short-term memory, which can capture the spatial features and temporal correlations among successive frames at the cost of very little computational resource.
Extensive experiments on three public datasets show that the proposed ESTINet can achieve faster speed than the competitors, while maintaining superior performance over the state-of-the-art methods. \href{https://github.com/HDCVLab/Enhanced-Spatio-Temporal-Interaction-Learning-for-Video-Deraining}{https://github.com/HDCVLab/Enhanced-Spatio-Temporal-Interaction-Learning-for-Video-Deraining}.
\end{abstract}

\begin{IEEEkeywords}
Video deraining; spatio-temporal learning; faster and better; ESTINet.
\end{IEEEkeywords}}

\maketitle

\IEEEdisplaynontitleabstractindextext

%
\IEEEpeerreviewmaketitle

\section{Introduction}
\label{introduction}

Images and videos captured by cameras in the outdoor scenarios often suffer from bad weather conditions. As one common condition, rain streaks cause a series of visibility degradations, seriously deteriorating the performance of outdoor vision-based systems. The goal of deraining is to remove those undesired rain streaks and recover sharp images from the input rainy versions. This is an active research topic in the computer vision field as it is vital to the robustness of modern intelligent systems.

\begin{figure}[t] 
  \centering
  {\includegraphics[width=0.99\linewidth]{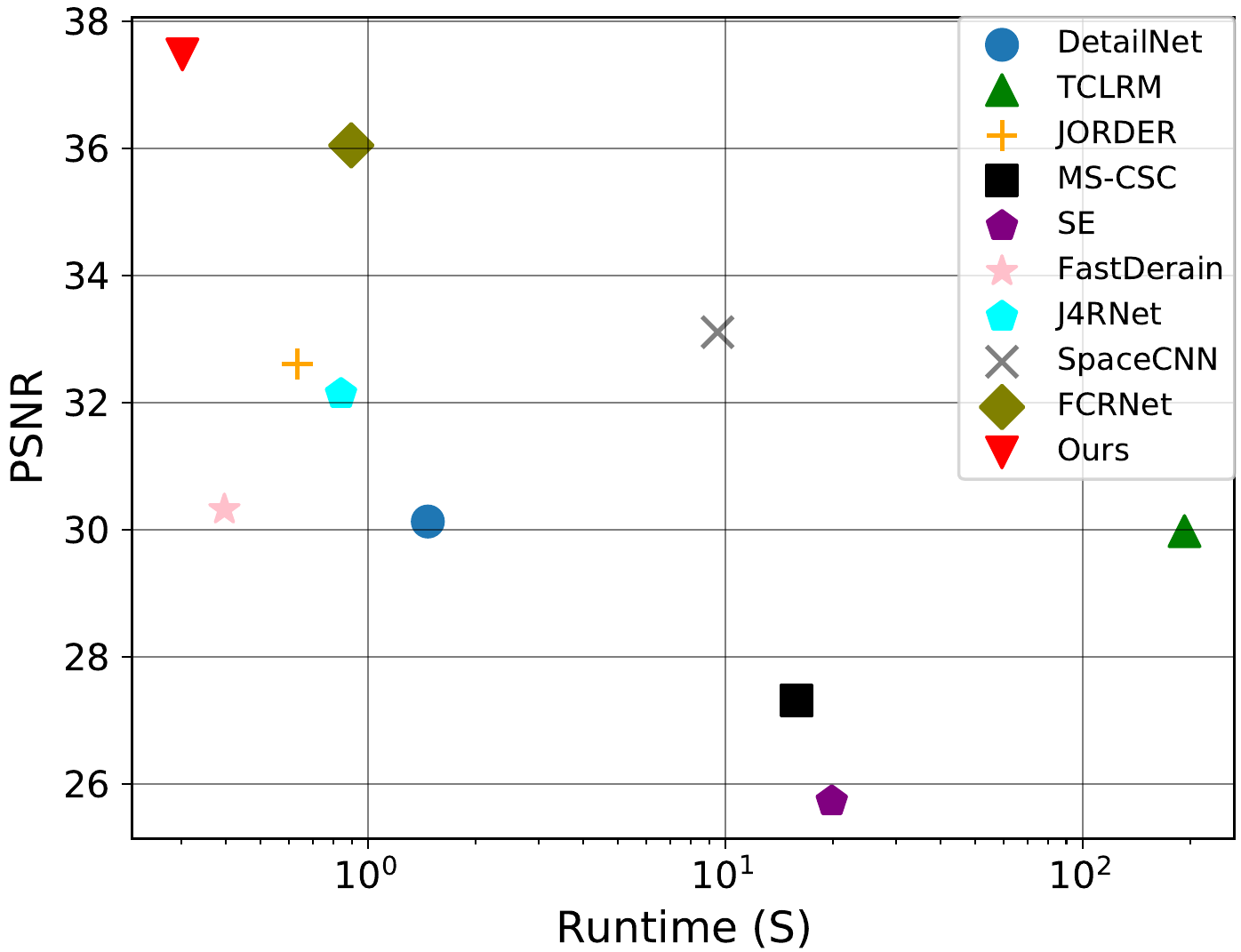}}
  \caption{The PSNR versus runtime of the state-of-the-art deep video deraining methods and our method on the NTURain dataset.} 
  \label{fig:speed_performance} 
\end{figure}

Based on different inputs, the deraining methods can be divided into two groups: image deraining and video deraining. In contrast to image deraining methods, which rely solely on the texture appearances of single frames, video deraining is a more challenging task as one has to consider how to model and exploit the inherent temporal correlations among continuing video frames. Moreover, several video deraining methods \cite{yang2019frame,liu2018erase,chen2018robust} achieve state-of-the-art performance but their speed is relatively slow. There also exists a method \cite{jiang2018fastderain} introducing fast video deraining models. However, the performance is far behind the current state-of-the-art methods. Therefore, the ideal approach of video deraining is to find an effective model to learn more powerful spatio-temporal features among successive frames with higher speed (Fig. \ref{fig:speed_performance}).

\begin{figure*}[t] 
  \centering 
  {\includegraphics[width=0.99\linewidth]{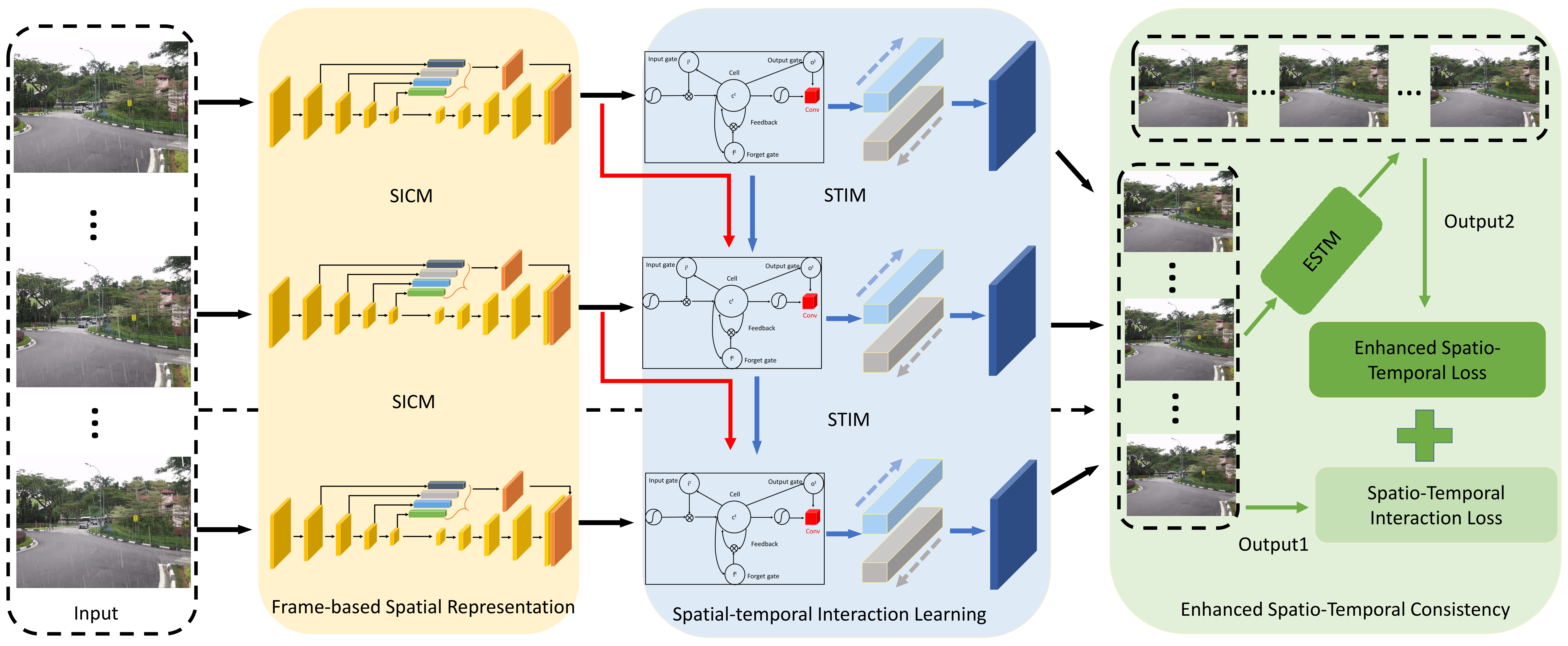}}
  \caption{ Our proposed Enhanced Spatio-Temporal Interaction Networks (\textit{ESTINet}). The input rainy frames are fed into SICM to extract the spatial cue, which is further forwarded into \textit{STIM} to extract spatio-temporal features. Finally, the proposed \textit{ESTM} takes the extracted features as input to capture the spatio-tempoal consistency and generate the final results. }
  \label{fig:overall_arc} 
\end{figure*}

In this paper, we propose an Enhanced Spatio-Temporal Integration Network (\textit{ESTINet}) to exploit the spatio-temporal information for rain streak removal. Fig. \ref{fig:overall_arc} illustrates the overall architecture of \textit{ESTINet}. It contains three parts: spatial information collection module (\textit{SICM}), spatio-temporal interaction module (\textit{STIM}), and enhanced spatio-temporal module (\textit{ESTM}). 

Considering that the spatial information plays an important role in video deraining, we firstly build an architecture called \textit{SICM} to directly extract high-level spatial features from the input rainy frames. Then the representations are fed into the second part, \textit{STIM}, to recover the coarsely derained frames. \textit{STIM} is a convolutional bidirectional long short-term memory (\textit{CBLSTM}) like architecture, called \textit{Interaction-CBLSTM}, which can directly make use of spatial features captured from the previous module. Therefore, it is a light-weighted module and mainly considers the temporal correlations to help remove rain streaks with a very little increase in the computational cost. Meanwhile, the loss calculated based on the output of \textit{STIM} also helps update the \textit{SICM} to extract more powerful spatial features. Moreover, different from traditional \textit{CBLSTM}, our \textit{Interaction-CBLSTM} (Fig. \ref{fig:stim}) architecture connects the features extracted from the last frame to the input and uses the convolutional operation to replace the \textit{tanh} function to adapt to different scales of input frames.
Finally, \textit{ESTM} takes the coarse deraining video as input and refines the temporal transformation with a 3D DenseNet-like architecture while preserving the realistic content information.

In summary, the contributions of this paper are three-fold:
\begin{itemize}

\item
\textbf{Framework level}: We construct a novel Enhanced Spatial-Temporal Interaction Network (\textit{ESTINet}) to extract better spatial-temporal information for video deraining. The proposed framework consists of three modules. The first and second modules are able to extract spatial information and temporal information, respectively. The last module is helpful to extract the enhanced spatio-temporal information. In this way, the proposed framework is capable of learning both spatial and temporal cues for video deraining.

\item
\textbf{Module level}: We develop a spatio-temporal interaction module of a convolutional bidirectional long short-term memory (\textit{CBLSTM}) like architecture. Different from traditional \textit{CBLSTM}, our \textit{Interaction-CBLSTM} architecture connects the features extracted from the last frame to the input and uses the convolutional operation to replace the \textit{tanh} function to adapt to different scales of input frames.

\item
\textbf{Performance}: Experiments on three public rainy video datasets show that the proposed \textit{ESTINet} achieves the state-of-the-art performance on video deraining. Meanwhile, in terms of speed, the \textit{ESTINet} also outperforms its counterparts.

\end{itemize}

\section{Related Work}



Single image deraining is a highly ill-posed problem, which aims to remove the rain from the background via analyzing only the visual information from a single image. In the recent decades, a set of models are proposed to recover the clean image from a rainy one, including local photo-metric, geometric, statistical properties of rain streaks \cite{garg2005does,zheng2013single,kim2013single,li2016rain} and deep learning methods \cite{li2019single,zhang2019image,fu2017clearing,fu2017removing,yang2017deep,zhang2018density,li2018recurrent,eigen2013restoring,qian2018attentive,zheng2019residual,wang2019spatial,wang2020rethinking,deng2019drd}.

In order to make use of the temporal corrections among video sequence frames, several video-based deraining methods are proposed and show a huge advantage for removing rain \cite{garg2004detection,barnum2010analysis,santhaseelan2012phase,santhaseelan2015utilizing,you2015adherent}. The early work focuses on capturing the temporal context and motion information via prior-based methods \cite{garg2004detection,garg2006photorealistic}. These kinds of methods model the rain streaks based on the photo-metric appearance of rain \cite{zhang2006rain,liu2009pixel,santhaseelan2015utilizing,brewer2008using,jiang2017novel} and propose learn-based models to address the problem of video deraining \cite{chen2013rain,tripathi2012video,kim2015video,wei2017should,ren2017video}. For example, Zhang \textit{et al.} \cite{zhang2006rain} combined temporal and chromatic properties to remove rain from video. Santhaseelan \textit{et al.} \cite{santhaseelan2015utilizing} and Barnum \textit{et al.} \cite{barnum2010analysis} removed rain streaks via extracting phase congruence features and Fourier domain features, respectively. Kim \textit{et al.} \cite{kim2015video} proposed a temporal correlation and low-rank matrix completion method to remove rain based on the observation that rain streaks cannot affect the optical flow estimation between frames.

Recently, many deep learning based methods have been proposed and brought significant changes to the video deraining \cite{li2018video,liu2018d3r,liu2018erase,chen2018robust,yang2019frame,yan2021self,yue2021semi}. Chen \textit{et al.} \cite{chen2018robust} firstly used a super-pixel segmentation scheme to decompose the image into depth consistent unites, and then restored clean video via a robust deep CNN. Liu \textit{et al.} presented a recurrent neural network to classify all pixels in rain frames, remove rain and reconstructed background details in \cite{liu2018erase}, and introduced a dynamic routing residue recurrent network to integrate their proposed hybrid rain model in \cite{liu2018d3r}. In order to make use of the additional degradation factors in the real world,  Yang \textit{et al.} \cite{yang2019frame} built a two-stage recurrent network to firstly capture motion information and then kept the motion consistency between frames to remove rain. There also exists self-learning deep video deraining method \cite{yang2020self}, which can learn how to remove rain without pairs of training samples. 

Although the above deep deraining methods achieve great success in video deraining, most of them focus on the performance and ignore the computational time. In this paper, we present a novel end-to-end video deraining method, which can improve the performance with higher speed.

\begin{figure*}[t] 
  \centering
  {\includegraphics[width=0.9\linewidth]{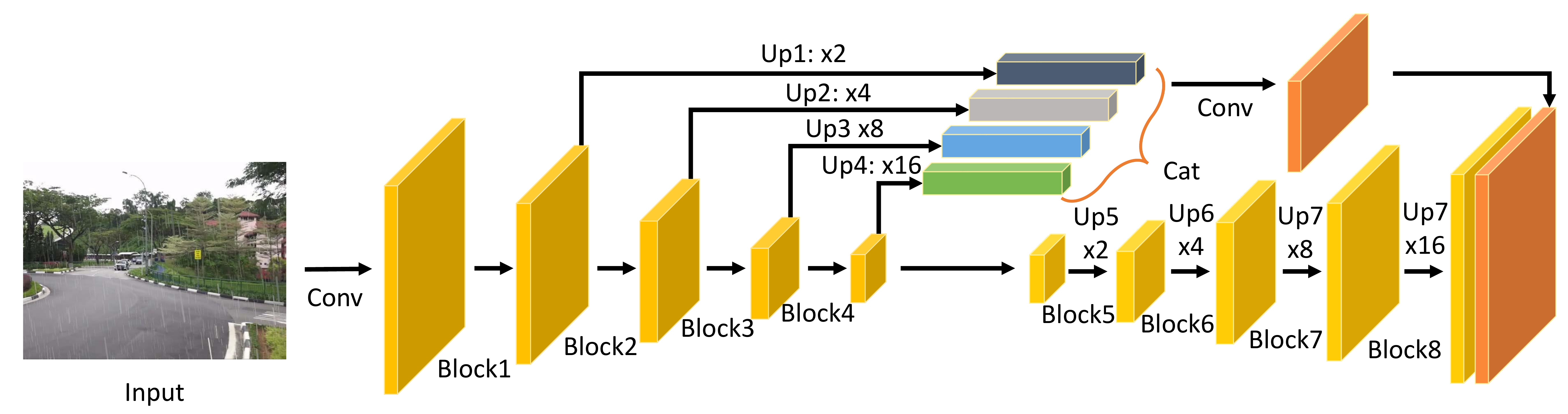}}
  \caption{The illustration of the ResNet-based Encoder-Decoder backbone (\textit{SICM}) to extract spatial representations from frames. The input is a single rainy frame, while the output is its spatial features. ``Up" means the upsampling operation.}
  \label{fig:sicm} 
\end{figure*}

\begin{figure*}[tb]
  \centering
  \subfigure[LSTM]{
    \label{}
    \includegraphics[width= 0.41\linewidth]{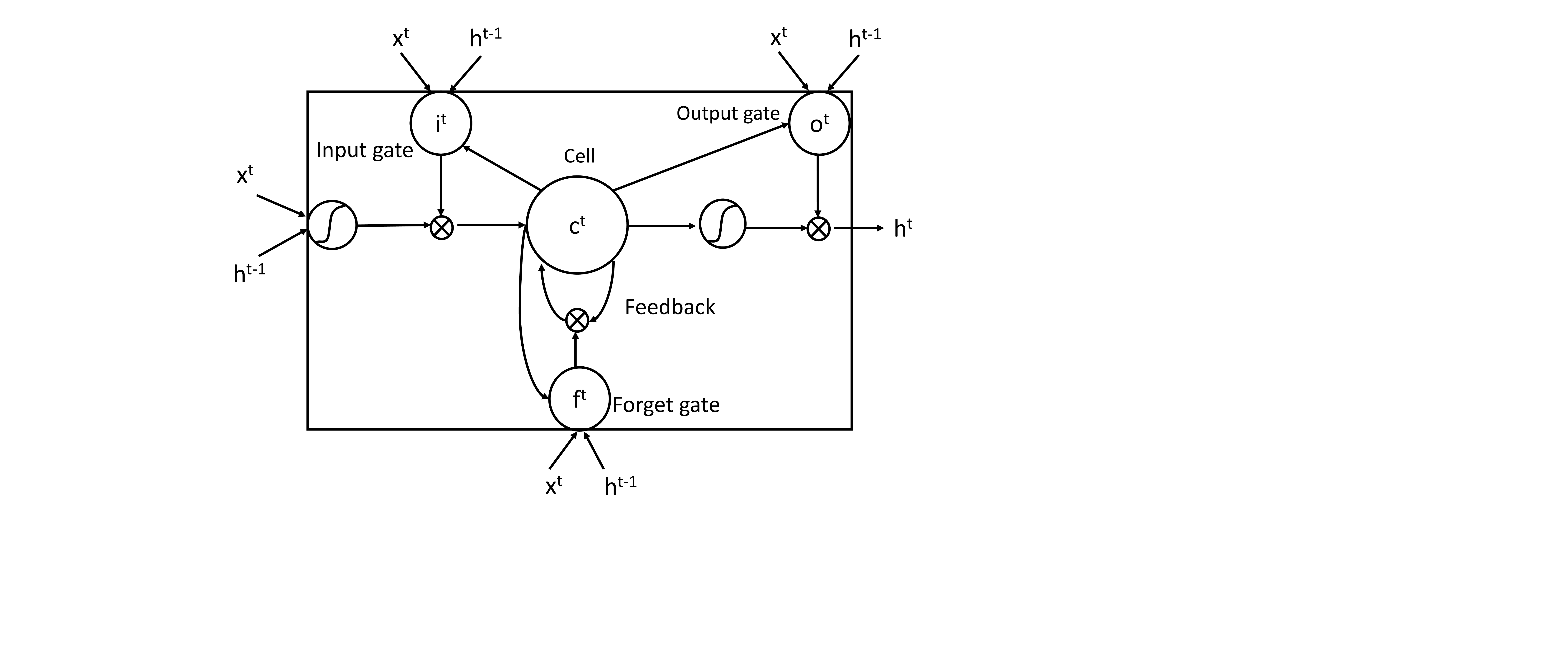}}
  \subfigure[Interaction-CBLSTM]{
    \label{}
    \includegraphics[width=0.56\linewidth]{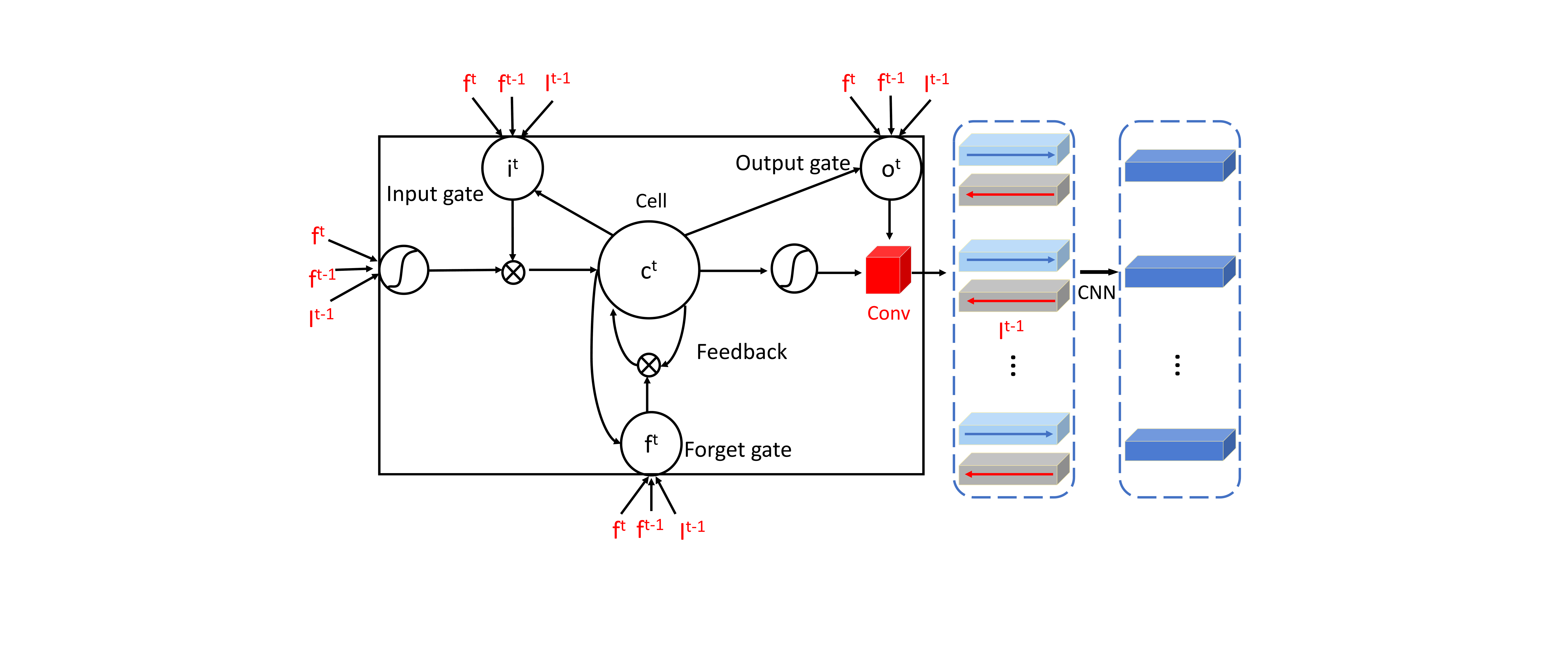}}
  \caption{ The comparison illustration between LSTM and the Interaction-CBLSTM backbone in \textit{STIM} to learn temporal correlations. The input ($f^{t}$) are the spatial features extracted from SICM illustrated in Fig. \ref{fig:sicm}. The output of ``Conv" is two kinds of spatio-temporal features which mean bidirectional sequences, which are further fed into a CNN to obtain the forward sequence. The blue and red arrows in the Interaction-CBLSTM mean the different orders of input frames.
  }
  \label{fig:stim} 
\end{figure*}

\section{ESTINet}
\label{sec:approach}

\subsection{Overall Architecture}
\label{sec:architecture}

The ultimate goal of our work is to remove the rain streaks and recover clean videos. In order to extract powerful spatio-temporal information efficiently, an Enhanced Spatio-Temporal Interaction Network, termed as \textit{ESTINet}, is proposed to extract features across both the spatial and temporal domains with a reduced computational cost. In this section, we will first introduce an \textit{SICM} architecture to extract a spatial representation from each input rainy frame in Sec. \ref{sec:spatial}. Then, the spatial representations are fed into our proposed \textit{STIM} to exploit the temporal information among successive frames (Sec. \ref{sec:temporal}). Finally, we build a 3D-DenseNet backbone, \textit{ESTM}, to enhance the spatial-temporal consistency in Sec. \ref{sec:spatio-temporal}. Fig. \ref{fig:overall_arc} shows the overall architecture of our proposed framework. During the training and testing stages, a rainy video, which can be regarded as multi frames, is divided into different groups and then fed into the proposed framework.

\subsection{Frame-based Spatial Representation}
\label{sec:spatial}

As shown in Fig. \ref{fig:sicm}, our \textit{SICM} is an Encoder-Decoder architecture. Both the encoder and decoder include one convolutional layer and four ResBlocks. The input are original RGB images. Following the input, the convolutional layer encodes the RGB images into feature maps with the same size as the original input. Then the four ResBlocks in the encoder employ four down-projection operations to decrease the resolution of the feature maps to their $1/16$. The decoder reconstructs clean images with original resolution via four up-projection operations. In order to fuse multi-scale features, there exists a multi-scale fusion module between the encoder and decoder. Specifically, the features maps extracted from the Block 2-5 are upsampled and then concatenated to the last layer of SICM. Among the SICM, we use a deep auto-encoder architecture for two reasons. First, the auto-encoder architecture is popular in the field of image restoration and has a powerful ability to extract spatial features. Second, the following STIM has fewer convolutional layers. The additional decoder in the SICM can help generate final clean frames. The number of SICM is same as the number of input frames. All SICMs share weights during the training stage.

Spatial features play an important role in the task of image restoration. Different from existing methods, which extract spatial features from a single frame, the proposed architecture directly learns a spatial representation from a video sequence for the following processing. In addition, we use a relatively light-weighted encoder-decoder architecture. In this way, the proposed model can process the input frames with a high speed. It can also be replaced with some other state-of-the-art backbones to improve the ability of spatial feature extraction.

\subsection{Spatial-Temporal Interaction Learning}
\label{sec:temporal}

After obtaining the spatial representation from the stack of input frames, we propose an \textit{STIM} to learn the temporal correlation between the continuous frames. The structure of \textit{STIM} is based on an LSTM model, which is shown in Fig. \ref{fig:stim}. The traditional LSTM can be formulated as follows:
\begin{equation}
f^{(t)} = \sigma(W^{(f)}x^{(t)}+W^{(f)}h^{(t-1)}+b^{(f)}), 
\end{equation}
\begin{equation}
i^{(t)} = \sigma(W^{(i)}x^{(t)}+W^{(i)}h^{(t-1)}+b^{(i)}), 
\end{equation}
\begin{equation}
\widetilde{C}^{(t)} = \tanh(W^{(C)}x^{(t)}+W^{(C)}h^{(t-1)}+b^{(C)}), 
\end{equation}
\begin{equation}
{C}^{(t)} = f^{(t)} \odot C^{(t-1)} +i^{(t)} \odot \widetilde{C}^{(t)}, 
\end{equation}
\begin{equation}
o^{(t)} = \sigma(U^{(o)}x^{(t)}+W^{(o)}h^{(t-1)}+b^{(o)}), 
\end{equation}
\begin{equation}
{h}^{(t)} = o^{(t)} \odot \tanh{C^{(t)}},
\end{equation}
where $U^{(\cdot)}$ and $W^{(\cdot)}$ are the input-to-hidden and hidden-to-hidden weight matrices, and $b^{(\cdot)}$ are bias vectors. $\sigma$ and $\odot$ are sigmoid activation function and point-wise multiplication, respectively. $x^{(t)}$ is the input of LSTM at time $t$. $h^{(t-1)}$ is the output of LSTM at time {t-1}.

Different from the traditional LSTM, which processes vectors, the proposed \textit{STIM} is modified based on the traditional LSTM to deal with video deraining. Firstly, the Hadamard product in LSTM is replaced with the convolution to address the 2D spatial representation extracted by \textit{SICM}. Secondly, we add the spatial representation of the last frame into the calculation of the forget gate $f^{(t)}$. Thirdly, we replace the hyperbolic tangent activation function with the convolution operation during the calculation of hidden state ${h}^{(t)}$ like ConvLSTM \cite{jiang2017predicting}, and add the bidirectional operation like bidirectional-LSTM \cite{graves2005bidirectional}. Our \textit{STIM} is formulated as:
\begin{equation}
\label{eq:stim}
f^{(t)} = \sigma(W^{(f)} \ast [f_x^{(t)}, f_x^{(t-1)}, h^{(t-1)}] +b^{(f)}), 
\end{equation}
\begin{equation}
i^{(t)} = \sigma(W^{(i)} \ast [f_x^{(t)}, f_x^{(t-1)}, h^{(t-1)}] +b^{(i)}), \end{equation}
\begin{equation}
\widetilde{C}^{(t)} = \tanh(W^{(C)} \ast [f_x^{(t)}, f_x^{(t-1)}, h^{(t-1)}] +b^{(C)}),
\end{equation}
\begin{equation}
{C}^{(t)} = f^{(t)} \odot C^{(t-1)} +i^{(t)} \odot \widetilde{C}^{(t)}, \end{equation}
\begin{equation}
o^{(t)} = \sigma(W^{(o)} \ast [f_x^{(t)}, f_x^{(t-1)}, h^{(t-1)}] + b^{(o)}), \end{equation}
\begin{equation}
{h}^{(t)} = Conv(o^{(t)}, \tanh{C^{(t)}}),
\end{equation}
\begin{equation}
{I_f}^{(t)} = Conv({h}^{(t)}, {h'}^{(t)}),
\end{equation}
where $\ast$ is the convolution operation. $f_x^{t}$ contains spatial feature maps extracted from frame $t$ by \textit{SICM}. We concatenate $f_x^{t}$ with the spatial feature maps $f_x^{t-1}$ extracted from the last frame $t-1$, and then feed them into \textit{STIM} to update the information. Then, the information from the output gate and updated memory cell is concatenated and fed into two convolutional layers to obtain the restoration results ${h}^{(t)}$. We can also obtain the other results ${h'}^{(t)}$ by reserving the order of frames. The results from two directions are finally fed into another two convolutional layers to obtain finer derained results ${I_f}^{(t)}$. Thanks to the LSTM architecture, the proposed Interaction-CBLSTM can benefit from all frames during the processing of rain removal.

\subsection{Enhanced Spatial-Temporal Consistency}
\label{sec:spatio-temporal}

The \textit{SICM} and \textit{STIM} work together to restore clean videos from input rainy versions. In order to enhance the spatio-temporal consistency and make full use of the correlations between continuous frames, we input the coarse results from \textit{STIM} into \textit{ESTM} to further improve the quality of the generated videos.

\begin{figure}[t] 
  \centering
  {\includegraphics[width=0.7\linewidth]{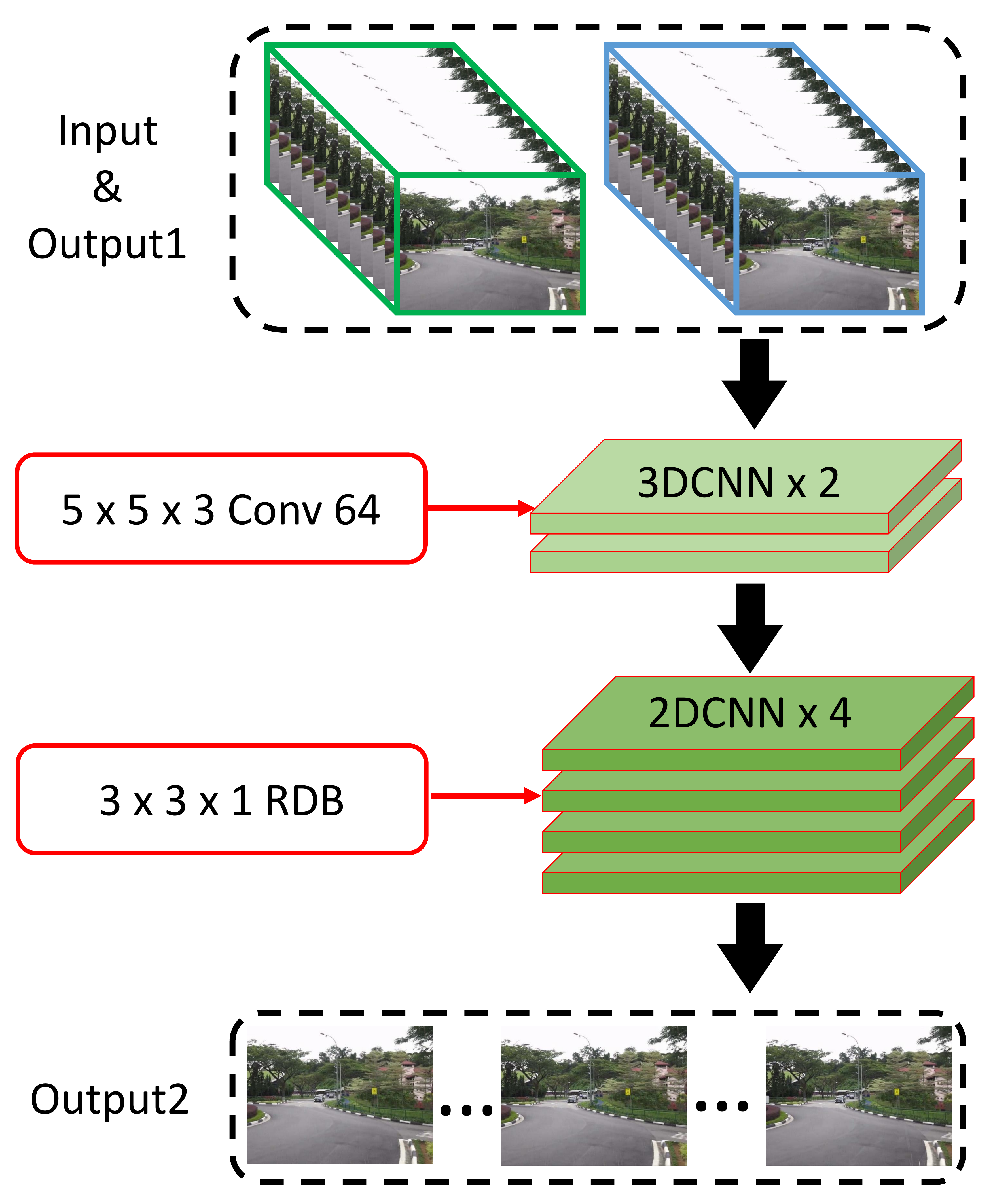}}
  \caption{The illustration of the Enhanced Spatio-Temporal Model (ESTM) to refine the deraining videos. ``Output1" represents the coarse deraining results, and ``Output2" indicates the refined results.}
  \label{fig:estm} 
\end{figure}

When training the \textit{SICM} and \textit{STIM}, we find it difficult to remove heavy rain while maintaining realistic content details. In other words, the \textit{SICM} and \textit{STIM} are helpful to remove most of the rain artifacts and restore coarse results, but may not be able to generate a better video and remove heavy rain. Therefore, we build a new architecture,  which is illustrated in Fig. \ref{fig:estm}.

Besides the ConvLSTM, which is able to capture the temporal correlations between continuous frames, 3D CNN is another popular architecture. In this paper, we apply 3D CNN in \textit{ESTM} to cover the shortage of the traditional LSTM and refine the deraining results. The coarse results and the original rainy frames are concatenated and fed into the \textit{ESTM}, which operates 3D convolutions via convolving 3D kernels on these frames. By doing so, the feature maps in convolutional layers can also capture the dynamic variations to help further remove rain and recover the details of images. Specially, we perform 3D convolution with kernel size of $3 \times 3 \time 6$ in the first and second convolutional layers to reduce the temporal dimension from five to one. In the following layers, we use the 2D convolution, RDB \cite{zhang2018residual}, to replace 3D operation as their temporal dimensions have already been decreased to one. The RDB aims to extract features via dense connected convolutional layers, which is similar to DenseNet \cite{huang2017densely}.

\subsection{Loss Functions}

In our work, we use two types of loss functions to train the proposed framework.

\textbf{Spatio-Temporal Interaction Loss}. The \textit{SICM} and \textit{STIM} are able to learn the spatial representations and temporal correlations from input frames. In order to help them interact with each other to recover coarse results, we apply the Mean Square Error (MSE) to calculate the spatio-temporal interaction loss, which is defined as:
\begin{equation}
{\mathcal{L}_{STI}} = \frac{1}{{WH}}\sum\limits_{x = 1}^W {\sum\limits_{y = 1}^H {{{(I_{x,y}^{clean} - G(I^{rainy})_{x,y})}^2}} }\, ,
\end{equation}
where $W$ and $H$ are the width and height of a frame, and $I_{x,y}^{clean}$ and $G(I^{rainy})_{x,y}$ correspond to the values of coarse derained frames and rainy frames at location $\left(x,y\right)$. Note that, as this loss measures the results from the \textit{SICM} and \textit{STIM}, which are dedicated to spatial and temporal domains, we call this loss spatio-temporal interaction loss.

\textbf{Enhanced Spatio-Temporal Loss}. Our proposed framework is a two-stage architecture. In order to drive our framework to generate finer derained frames, we introduce another loss function to refine the coarse results. During the training stage, the parameter of \textit{ESTM} is updated based on the Enhanced Spatio-Temporal loss to further remove rain and recover clean images. The loss function can be represented as:
\begin{equation}
{\mathcal{L}_{EST}} = \frac{1}{{WH}}\sum\limits_{x = 1}^W {\sum\limits_{y = 1}^H {{{(I_{x,y}^{clean} - G(I^{rainy}, I^{derained})_{x,y})}^2}} }\, ,
\end{equation}
where $I^{derained}$ collects the coarse derained frames generated from the \textit{STIM}.
This loss is used to assess the enhanced results regarding the ground truth, so we term it the enhanced spatio-temporal loss.

\textbf{Balance of Different Loss Functions}. In the training stage, the above two loss functions are combined as:
\begin{equation}
{\mathcal{L}_{final}} = {\mathcal{L}_{STI}} + \alpha \cdot {\mathcal{L}_{EST}},
\end{equation}
where $\alpha$ is a hyper-parameter to balance the two loss functions. In this paper, we set it as $1$.

\begin{figure*}[t] 
  \centering
  {\includegraphics[width=0.95\linewidth]{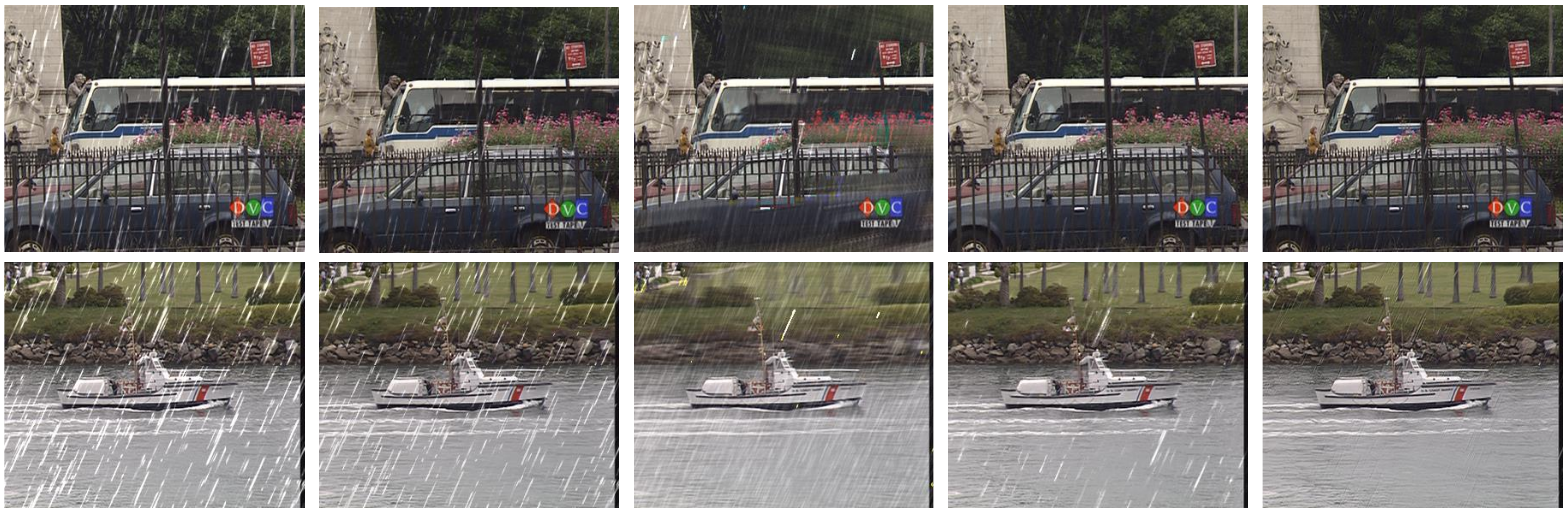}}
  \caption{{\bf Exemplar results on the RainSynLight25 dataset}. From left to right: input, results of \cite{jiang2018fastderain}, \cite{wei2017should},  \cite{chen2018robust} and ours. All results are attained without alignment. Best viewed in color.}
  \label{fig:rainsynlight} 
\end{figure*}


\begin{figure*}[ht] 
  \centering
  {\includegraphics[width=0.95\linewidth]{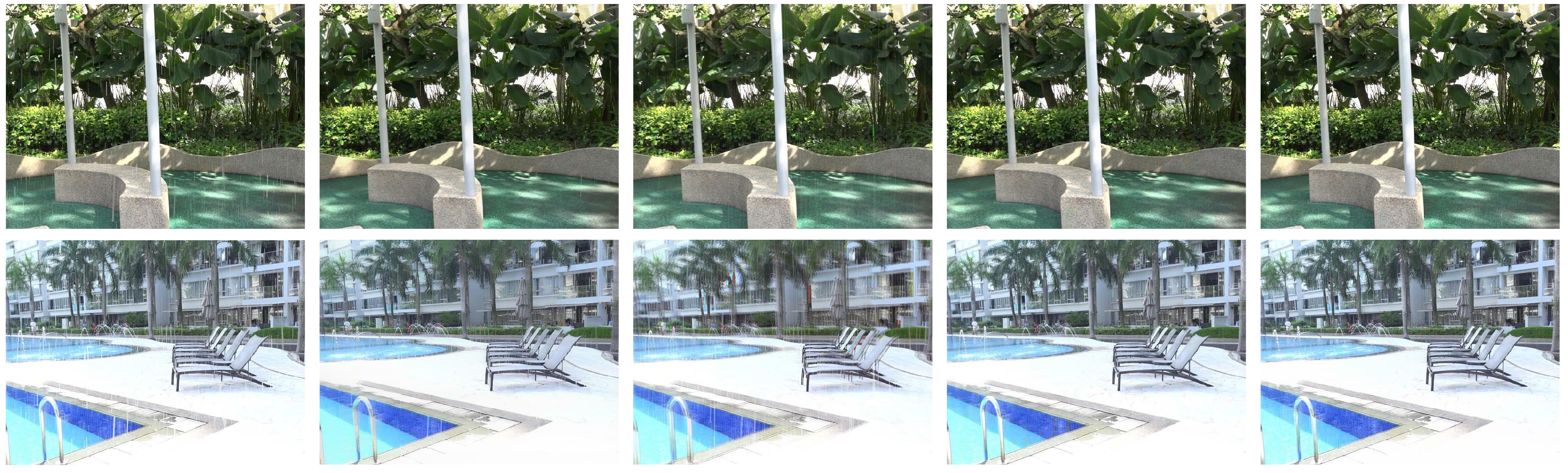}}
  \caption{{\bf Exemplar results on the NTURain dataset}. From left to right: input, results of \cite{jiang2018fastderain}, \cite{wei2017should}, \cite{chen2018robust} and ours. All results are attained without alignment. Best viewed in color.}
  \label{fig:nturain} 
\end{figure*}

\begin{figure*}[ht] 
  \centering
  {\includegraphics[width=0.95\linewidth]{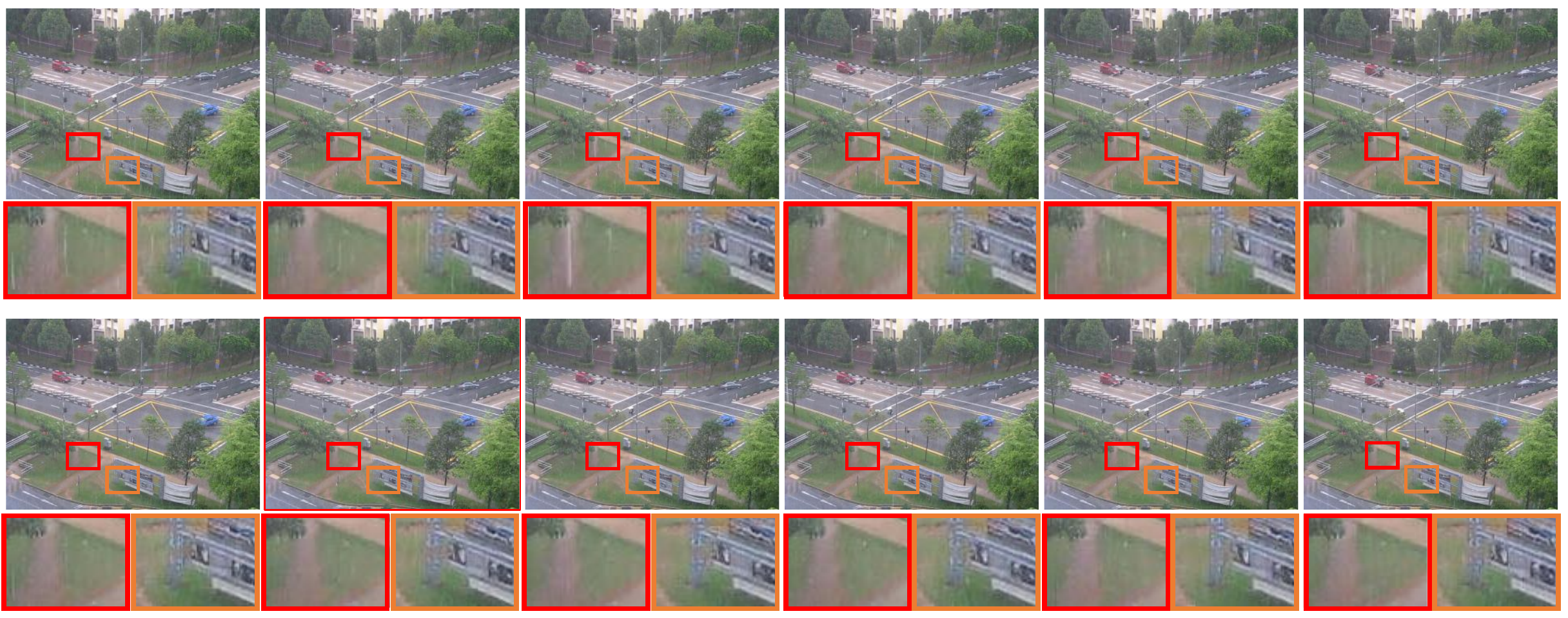}}
  \caption{{\bf Deraining results on the real-world rainy sequences}. The top and bottom rows are the input sequences and the output sequences from out proposed model, respectively. Best viewed in color.}
  \label{fig:real} 
\end{figure*}

\begin{table*}[t]
  \centering 
  \scriptsize
    \caption{Performance comparison with state-of-the-art methods on the RainSynLight25, RainSynHeavy25 and NTURain datasets.}
    \begin{tabular}{l | c | c c c c c c c c c | c }
    \toprule
    Dataset &  Metric & DetailNet & TCLRM & JORDER & MS-CSC & SE & FastDerain & J4RNet & SPAC & FCRNet & \textbf{Ours} \\
    \hline 
    \multirow{2}{*}{\textit{NTURain}}   & PSNR & 30.13& 29.98& 32.61& 27.31 & 25.73 & 30.32 & 32.14 & 33.11 & 36.05 & \textbf{37.48} \\
      & SSIM & 0.9220 & 0.9199 & 0.9482 & 0.7870 & 0.7614 & 0.9262 & 0.9480 & 0.9474 & 0.9676 & \textbf{0.9700} \\
    \hline
    \multirow{2}{*}{\textit{RainSynLight25}} & PSNR & 25.72 & 28.77 & 30.37 & 25.58 & 26.56 & 29.42 & 32.96 & 32.78& 35.80 & \textbf{36.12} \\
     & SSIM & 0.8572 & 0.8693 & 0.9235 & 0.8089 & 0.8006 & 0.8683 & 0.9434 & 0.9239 & 0.9622 & \textbf{0.9631} \\
    \hline
    \multirow{2}{*}{\textit{RainSynHeavy25}}  & PSNR & 16.50 & 17.31 & 20.20 & 16.96 & 16.76 & 19.25 & 24.13 & 21.21 & 27.72 & \textbf{28.48}  \\
     & SSIM & 0.5441 & 0.4956 & 0.6335 & 0.5049 & 0.5293 & 0.5385 & 0.7163& 0.5854& 0.8239& \textbf{0.8242} \\
    \bottomrule
    \end{tabular}%
    \label{table:sota}
\end{table*}%

\section{Experiments}

We test our approach on three widely used public datasets, NTURain \cite{chen2018robust}, RainSynLight25, and RainSynComplex25 \cite{liu2018d3r}, which are introduced firstly in Sec. \ref{dataset}. Then we introduce the implementation details of our framework in Sec. \ref{implementation} and
compare our method with the state-of-the-art methods in Sec. \ref{comparison}. An ablation study is conducted to show the effectiveness of its different components in Sec. \ref{ablation}. An efficiency analysis is reported subsequently in Sec. \ref{exp:efficiency}.

\subsection{Datasets}
\label{dataset}

\textbf{NTURain}. This dataset is created by Chen \textit{et al.} \cite{chen2018robust}. The images are taken by a camera with slow and fast movements. The training contains $24$ rainy sequences and their corresponding clean versions, while the testing set contains $8$ pairs of sequences. In addition, it also provides seven real-world rainy videos.

\textbf{RainSynLight25}. It contains $190$ pairs of RGB rainy and clean sequences for training, and $27$ pairs for testing. The sharp images are from CIF testing sequences, HDTV sequences, and HEVC standard testing sequences. Via adding rain streaks generated by the probabilistic model \cite{garg2006photorealistic}, the corresponding rainy images are obtained.

\textbf{RainSynHeavy25}. This dataset is similar to the dataset of RainSynLight25. The main difference is that the rain streaks in rainy images are generated by the probabilistic model, sharp line streaks, and sparkle noises. Therefore, they are heavier than those in the RainSynLight25 dataset.

\subsection{Implementation Details}
\label{implementation}

The weights of networks in our framework are initialized via a Gaussian distribution with zero mean and a standard deviation of $0.01$. Models are updated after learning a mini-batch of size $8$ in each iteration. We also crop patches of size $224 \times 224$ from images, and randomly flip frames horizontally to augment the training set. During the training stage, we first train the \textit{SICM} and \textit{STIM} without the \textit{ESTM} module, and then update all weights of them. The learning rate is set as a value of $10^{-4}$ and reduces to $10^{-6}$ after the training loss gets converged. For evaluation, we employ PSNR and SSIM as metrics.

\begin{table*}[tb]
  \centering 
  \footnotesize
    \caption{Speed comparison with current methods. The numbers are in seconds. The parameters and FLOPs of deep deraining methods are also provided.}
    \begin{tabular}{l | c c c c c c c c c | c }
    \toprule
    Method &  DetailNet & TCLRM & JORDER & MS-CSC & SE & FastDerain & J4RNet & SPAC & FCRNet & \textbf{Ours} \\
    \hline 
    Speed & 1.4698 & 192.7007 & 0.6329 & 15.7957& 19.8516 & 0.3962& 0.8401& 9.5075 & 0.8974 & \textbf{0.3122} \\
    Param & 0.7M & - & 0.5M & - & -  & - & 0.6M & -  & 0.6M  & \textbf{0.4M} \\
    FLOPs & 3.6e11 & - & 2.7e11 & - & -  & - & 2.9e11 & -  & 3.0e11  & \textbf{1.5e11} \\
    \bottomrule
    \end{tabular}%
    \label{table:speed}
\end{table*}%

\subsection{Comparison with Existing Methods}
\label{comparison}

In this section, we compare the performance of our proposed framework with several state-of-the-art deraining methods on the above three widely used datasets. Among these methods, stochastic encoding (SE) \cite{starik2003simulation}, temporal correlation and low-rank matrix completion (TCLRM) \cite{kim2015video}, FastDerain \cite{jiang2018fastderain}, joint recurrent rain removal and reconstruction (J4RNet) \cite{liu2018erase} and superpixel alignment, compensation CNN (SPAC), and frame-consistent recurrent network (FCRNet) \cite{yang2019frame} are video-based deraining methods, joint rain detection and removal (JORDER) \cite{yang2017deep}, deep detail network (DetailNet) \cite{fu2017removing}, J4RNet \cite{liu2018erase}, SPAC \cite{chen2018robust} , and FCRNet \cite{yang2019frame} are deep deraining methods. Table \ref{table:sota} shows the quantitative comparison results between our method and the existing deraining methods. Before our work, FCRNet achieves the state-of-the-art performance on three public video deraining datasets. Our method further improves over the FCRNet method and obtains the best performance, in terms of both PSNR and SSIM values. This indicates that our framework achieves better feature representations due to the learned spatio-temporal interactions.
To give an intuitive view of how ours and these compared methods perform, 
Fig. \ref{fig:rainsynlight} and Fig. \ref{fig:nturain} show the exemplar visual results on the datasets of RainSynLight25 and NTURain. The qualitative comparison results also evidently verify that our method achieves better performance than the existing ones. 
In addition, we also show the performance of our approach in real-world scenarios. Taking a real-world rainy video from the NTURain dataset, we process this video by our method to remove the rain, and the result frames are shown in Fig. \ref{fig:real}. The rain is successfully removed to some extent.



\subsection{Ablation Study}
\label{ablation}


\begin{table}
  \centering 
    \caption{Performance comparison of different architectures on the NTURain dataset.}
    \begin{tabular}{l | c c }
    \toprule
    Methods &  PSNR & SSIM \\
    \hline 
    \textit{SICM  + 2DCNN} & 35.44 & 0.9562 \\
    \hline
    \textit{SICM + STIM (\#2)}  & 36.61 & 0.9668  \\
    \textit{SICM + STIM (\#3)}  & 36.93 & 0.9677 \\
    \textit{SICM + STIM (\#4)}  & 37.16 & 0.9682 \\
    \textit{SICM + STIM (\#5)}  & 37.28 & 0.9693 \\
    \hline
     \textit{SICM + ConvLSTM + ESTM}  & 36.43 & 0.9656  \\
     \textit{SICM + B-ConvLSTM + ESTM} & 36.76 & 0.9671  \\
    \textbf{SICM + STIM (\#5) + ESTM}  & \textbf{37.48} & \textbf{0.9700}  \\
    \textit{SICM + STIM (\#5) + 2DCNN}  & 37.35 & 0.9695  \\
    \bottomrule
    \end{tabular}%
    \label{table:ablation}
\end{table}%

The proposed \textit{STIM} has the advantage of capturing temporal correlations from successive frames and helping update the \textit{SICM} to learn better spatial representations. The \textit{ESTM} is able to learn enhanced spatio-temporal representations via making use of the coarse derained images and the 3D Convolution to refine the results. In order to verify its effectiveness, we develop nine variant networks: \textit{SICM  + 2DCNN}, \textit{SICM + STIM (\#2)}, \textit{SICM + STIM (\#3)}, \textit{SICM + STIM (\#4)}, \textit{SICM + STIM (\#5)}, \textit{SICM + ConvLSTM + ESTM}, \textit{SICM + B-ConvLSTM + ESTM}, \textit{SICM + STIM (\#5) + ESTM}, and \textit{SICM + STIM (\#5) + 2DCNN}. \textit{SICM  + 2DCNN} is a baseline method, which replaces the STIM with three ordinary convolutional layers. The input to \textit{SICM  + 2DCNN} is a single frame, so it does not take into consideration of the temporal information among the consecutive frames. In order to show that how the number of input frames influences the performance of our proposed model, we compare the values of PSNR and SSIM of models regarding different numbers of input rainy frames. Specially, the number $n$ in \textit{SICM + STIM (\#n)} represents the number of consecutive frames.

The quantitative results are shown in Table \ref{table:ablation}. Specifically, by learning temporal information from consecutive frames, all the variants of \textit{SICM + STIM (\#n)} outperform the plain model \textit{SICM  + 2DCNN}, which verifies the usefulness of the temporal information. And with the increase of input frames, better performance is achieved. The \textit{SICM + STIM (\#5) + ESTM} achieves better performance than \textit{SICM + ConvLSTM + ESTM} and \textit{SICM + B-ConvLSTM + ESTM}, which shows the effectiveness of the STIM module. The \textit{SICM + STIM (\#5) + ESTM} achieves better performance than \textit{SICM + CBLSTM (\#5) + 2DCNN}, showing the effectiveness of the 3D convolution in ESTM. By considering the spatio-temporal interaction, our full method \textit{SICM + STIM (\#5) + ESTM} achieves additional gains. 

\subsection{Efficiency Analysis}
\label{exp:efficiency}


Table \ref{table:speed} shows the speed of the state-of-the-art deraining methods. J4RNet, FCRNet and our proposed methods are based on the PyTorch framework, while other methods are implemented based on Matlab. We evaluate the speed on the NTURain dataset on an ordinary platform. The platform is a normal desktop PC with GeForce GTX 1080 Ti GPU, which is the same as \cite{yang2019frame}. Therefore, we directly quote some related results from \cite{yang2019frame}. Our proposed method is significantly faster than other state-of-the-art methods, including the FastDeRain method.

\subsection{Discussion}
\label{exp:discussion}

The proposed framework achieves better performance for three reasons. The multi-scale SICM is able to extract good spatial information, the STIM  is able to extract useful temporal information, and the ESTM is able to enhance the spatial-temporal consistency and make good use of correlations between continuous frames. We use LSTM to replace 3D convolution in STIM because it is more flexible than 3D convolution. The reconstruction model in SICM is able to help STIM to generate derained images for calculating loss functions.

\section{Conclusion}

In this paper, we proposed a novel end-to-end framework to address the problem of video deraining by a \textbf{faster} scheme with \textbf{better} quantitative and qualitative results. To obtain the spatial representation, \textit{i.e.}, a ResNet-based architecture, \textit{SICM} is built to directly extract spatial features from a stack of input frames. The representations are then fed into a well-designed Interaction-\textit{CBLSTM} architecture, \textit{STIM}, to capture the temporal correlations. In the training stage, the proposed \textit{SICM} and \textit{STIM} interact with each other to capture the spatial information and temporal correlations between consecutive frames to obtain coarse results, which are fed into a 3D-DenseNet based architecture, \textit{ESTM}, to enhance the performance of rain removal and obtain finer results. The extensive experiments have verified that the proposed framework outperforms the state-of-the-art methods in terms of both quality and speed.


\bibliographystyle{IEEEtran}
\bibliography{egbib_new}

\begin{thebibliography}{10}
\providecommand{\url}[1]{#1}
\csname url@samestyle\endcsname
\providecommand{\newblock}{\relax}
\providecommand{\bibinfo}[2]{#2}
\providecommand{\BIBentrySTDinterwordspacing}{\spaceskip=0pt\relax}
\providecommand{\BIBentryALTinterwordstretchfactor}{4}
\providecommand{\BIBentryALTinterwordspacing}{\spaceskip=\fontdimen2\font plus
\BIBentryALTinterwordstretchfactor\fontdimen3\font minus
  \fontdimen4\font\relax}
\providecommand{\BIBforeignlanguage}[2]{{%
\expandafter\ifx\csname l@#1\endcsname\relax
\typeout{** WARNING: IEEEtran.bst: No hyphenation pattern has been}%
\typeout{** loaded for the language `#1'. Using the pattern for}%
\typeout{** the default language instead.}%
\else
\language=\csname l@#1\endcsname
\fi
#2}}
\providecommand{\BIBdecl}{\relax}
\BIBdecl

\bibitem{yang2019frame}
W.~Yang, J.~Liu, and J.~Feng, ``Frame-consistent recurrent video deraining with
  dual-level flow,'' in \emph{Proceedings of the IEEE Conference on Computer
  Vision and Pattern Recognition}, 2019, pp. 1661--1670.

\bibitem{liu2018erase}
J.~Liu, W.~Yang, S.~Yang, and Z.~Guo, ``Erase or fill? deep joint recurrent
  rain removal and reconstruction in videos,'' in \emph{Proceedings of the IEEE
  Conference on Computer Vision and Pattern Recognition}, 2018, pp. 3233--3242.

\bibitem{chen2018robust}
J.~Chen, C.-H. Tan, J.~Hou, L.-P. Chau, and H.~Li, ``Robust video content
  alignment and compensation for rain removal in a cnn framework,'' in
  \emph{Proceedings of the IEEE Conference on Computer Vision and Pattern
  Recognition}, 2018, pp. 6286--6295.

\bibitem{jiang2018fastderain}
T.-X. Jiang, T.-Z. Huang, X.-L. Zhao, L.-J. Deng, and Y.~Wang, ``Fastderain: A
  novel video rain streak removal method using directional gradient priors,''
  \emph{IEEE Transactions on Image Processing}, vol.~28, no.~4, pp. 2089--2102,
  2018.

\bibitem{garg2005does}
K.~Garg and S.~K. Nayar, ``When does a camera see rain?'' in \emph{Tenth IEEE
  International Conference on Computer Vision (ICCV'05) Volume 1},
  vol.~2.\hskip 1em plus 0.5em minus 0.4em\relax IEEE, 2005, pp. 1067--1074.

\bibitem{zheng2013single}
X.~Zheng, Y.~Liao, W.~Guo, X.~Fu, and X.~Ding, ``Single-image-based rain and
  snow removal using multi-guided filter,'' in \emph{International Conference
  on Neural Information Processing}.\hskip 1em plus 0.5em minus 0.4em\relax
  Springer, 2013, pp. 258--265.

\bibitem{kim2013single}
J.-H. Kim, C.~Lee, J.-Y. Sim, and C.-S. Kim, ``Single-image deraining using an
  adaptive nonlocal means filter,'' in \emph{2013 IEEE International Conference
  on Image Processing}.\hskip 1em plus 0.5em minus 0.4em\relax IEEE, 2013, pp.
  914--917.

\bibitem{li2016rain}
Y.~Li, R.~T. Tan, X.~Guo, J.~Lu, and M.~S. Brown, ``Rain streak removal using
  layer priors,'' in \emph{Proceedings of the IEEE conference on computer
  vision and pattern recognition}, 2016, pp. 2736--2744.

\bibitem{li2019single}
S.~Li, I.~B. Araujo, W.~Ren, Z.~Wang, E.~K. Tokuda, R.~H. Junior,
  R.~Cesar-Junior, J.~Zhang, X.~Guo, and X.~Cao, ``Single image deraining: A
  comprehensive benchmark analysis,'' in \emph{Proceedings of the IEEE
  Conference on Computer Vision and Pattern Recognition (CVPR)}, 2019.

\bibitem{zhang2019image}
H.~Zhang, V.~Sindagi, and V.~M. Patel, ``Image de-raining using a conditional
  generative adversarial network,'' \emph{IEEE Transactions on Circuits and
  Systems for Video Technology (TCSVT)}, 2019.

\bibitem{fu2017clearing}
X.~Fu, J.~Huang, X.~Ding, Y.~Liao, and J.~Paisley, ``Clearing the skies: A deep
  network architecture for single-image rain removal,'' \emph{IEEE Transactions
  on Image Processing (TIP)}, 2017.

\bibitem{fu2017removing}
X.~Fu, J.~Huang, D.~Zeng, Y.~Huang, X.~Ding, and J.~Paisley, ``Removing rain
  from single images via a deep detail network,'' in \emph{Proceedings of the
  IEEE Conference on Computer Vision and Pattern Recognition (CVPR)}, 2017.

\bibitem{yang2017deep}
W.~Yang, R.~T. Tan, J.~Feng, J.~Liu, Z.~Guo, and S.~Yan, ``Deep joint rain
  detection and removal from a single image,'' in \emph{Proceedings of the IEEE
  Conference on Computer Vision and Pattern Recognition (CVPR)}, 2017.

\bibitem{zhang2018density}
H.~Zhang and V.~M. Patel, ``Density-aware single image de-raining using a
  multi-stream dense network,'' in \emph{Proceedings of the IEEE Conference on
  Computer Vision and Pattern Recognition (CVPR)}, 2018.

\bibitem{li2018recurrent}
X.~Li, J.~Wu, Z.~Lin, H.~Liu, and H.~Zha, ``Recurrent squeeze-and-excitation
  context aggregation net for single image deraining,'' in \emph{European
  Conference on Computer Vision (ECCV)}, 2018.

\bibitem{eigen2013restoring}
D.~Eigen, D.~Krishnan, and R.~Fergus, ``Restoring an image taken through a
  window covered with dirt or rain,'' in \emph{Proceedings of the IEEE
  International Conference on Computer Vision (ICCV)}, 2013.

\bibitem{qian2018attentive}
R.~Qian, R.~T. Tan, W.~Yang, J.~Su, and J.~Liu, ``Attentive generative
  adversarial network for raindrop removal from a single image,'' in
  \emph{Proceedings of the IEEE Conference on Computer Vision and Pattern
  Recognition (CVPR)}, 2018.

\bibitem{zheng2019residual}
Y.~Zheng, X.~Yu, M.~Liu, and S.~Zhang, ``Residual multiscale based single image
  deraining.'' in \emph{British Machine Vision Conference (BMVC)}, 2019.

\bibitem{wang2019spatial}
T.~Wang, X.~Yang, K.~Xu, S.~Chen, Q.~Zhang, and R.~W. Lau, ``Spatial attentive
  single-image deraining with a high quality real rain dataset,'' in
  \emph{Proceedings of the IEEE/CVF Conference on Computer Vision and Pattern
  Recognition}, 2019, pp. 12\,270--12\,279.

\bibitem{wang2020rethinking}
Y.~Wang, Y.~Song, C.~Ma, and B.~Zeng, ``Rethinking image deraining via rain
  streaks and vapors,'' in \emph{European Conference on Computer Vision}.\hskip
  1em plus 0.5em minus 0.4em\relax Springer, 2020, pp. 367--382.

\bibitem{deng2019drd}
S.~Deng, M.~Wei, J.~Wang, L.~Liang, H.~Xie, and M.~Wang, ``Drd-net:
  Detail-recovery image deraining via context aggregation networks,''
  \emph{arXiv preprint arXiv:1908.10267}, 2019.

\bibitem{garg2004detection}
K.~Garg and S.~K. Nayar, ``Detection and removal of rain from videos,'' in
  \emph{Proceedings of the 2004 IEEE Computer Society Conference on Computer
  Vision and Pattern Recognition, 2004. CVPR 2004.}, vol.~1.\hskip 1em plus
  0.5em minus 0.4em\relax IEEE, 2004, pp. I--I.

\bibitem{barnum2010analysis}
P.~C. Barnum, S.~Narasimhan, and T.~Kanade, ``Analysis of rain and snow in
  frequency space,'' \emph{International journal of computer vision}, vol.~86,
  no. 2-3, p. 256, 2010.

\bibitem{santhaseelan2012phase}
V.~Santhaseelan and V.~K. Asari, ``A phase space approach for detection and
  removal of rain in video,'' in \emph{Intelligent Robots and Computer Vision
  XXIX: Algorithms and Techniques}, vol. 8301.\hskip 1em plus 0.5em minus
  0.4em\relax International Society for Optics and Photonics, 2012, p. 830114.

\bibitem{santhaseelan2015utilizing}
------, ``Utilizing local phase information to remove rain from video,''
  \emph{International Journal of Computer Vision}, vol. 112, no.~1, pp. 71--89,
  2015.

\bibitem{you2015adherent}
S.~You, R.~T. Tan, R.~Kawakami, Y.~Mukaigawa, and K.~Ikeuchi, ``Adherent
  raindrop modeling, detectionand removal in video,'' \emph{IEEE transactions
  on pattern analysis and machine intelligence}, vol.~38, no.~9, pp.
  1721--1733, 2015.

\bibitem{garg2006photorealistic}
K.~Garg and S.~K. Nayar, ``Photorealistic rendering of rain streaks,'' in
  \emph{ACM Transactions on Graphics (TOG)}, 2006.

\bibitem{zhang2006rain}
X.~Zhang, H.~Li, Y.~Qi, W.~K. Leow, and T.~K. Ng, ``Rain removal in video by
  combining temporal and chromatic properties,'' in \emph{IEEE International
  Conference on Multimedia and Expo (ICME)}, 2006.

\bibitem{liu2009pixel}
P.~Liu, J.~Xu, J.~Liu, and X.~Tang, ``Pixel based temporal analysis using
  chromatic property for removing rain from videos.'' \emph{Computer and
  Information Science}, 2009.

\bibitem{brewer2008using}
N.~Brewer and N.~Liu, ``Using the shape characteristics of rain to identify and
  remove rain from video,'' in \emph{Joint IAPR International Workshops on
  Statistical Techniques in Pattern Recognition (SPR) and Structural and
  Syntactic Pattern Recognition (SSPR)}, 2008.

\bibitem{jiang2017novel}
T.-X. Jiang, T.-Z. Huang, X.-L. Zhao, L.-J. Deng, and Y.~Wang, ``A novel
  tensor-based video rain streaks removal approach via utilizing
  discriminatively intrinsic priors,'' in \emph{Proceedings of the IEEE
  Conference on Computer Vision and Pattern Recognition (CVPR)}, 2017.

\bibitem{chen2013rain}
J.~Chen and L.-P. Chau, ``A rain pixel recovery algorithm for videos with
  highly dynamic scenes,'' \emph{IEEE Transactions on Image Processing (TIP)},
  2013.

\bibitem{tripathi2012video}
A.~Tripathi and S.~Mukhopadhyay, ``Video post processing: low-latency
  spatiotemporal approach for detection and removal of rain,'' \emph{IET Image
  Processing}, 2012.

\bibitem{kim2015video}
J.-H. Kim, J.-Y. Sim, and C.-S. Kim, ``Video deraining and desnowing using
  temporal correlation and low-rank matrix completion,'' \emph{IEEE
  Transactions on Image Processing (TIP)}, 2015.

\bibitem{wei2017should}
W.~Wei, L.~Yi, Q.~Xie, Q.~Zhao, D.~Meng, and Z.~Xu, ``Should we encode rain
  streaks in video as deterministic or stochastic?'' in \emph{Proceedings of
  the IEEE International Conference on Computer Vision (ICCV)}, 2017.

\bibitem{ren2017video}
W.~Ren, J.~Tian, Z.~Han, A.~Chan, and Y.~Tang, ``Video desnowing and deraining
  based on matrix decomposition,'' in \emph{Proceedings of the IEEE Conference
  on Computer Vision and Pattern Recognition (CVPR)}, 2017.

\bibitem{li2018video}
M.~Li, Q.~Xie, Q.~Zhao, W.~Wei, S.~Gu, J.~Tao, and D.~Meng, ``Video rain streak
  removal by multiscale convolutional sparse coding,'' in \emph{Proceedings of
  the IEEE Conference on Computer Vision and Pattern Recognition}, 2018, pp.
  6644--6653.

\bibitem{liu2018d3r}
J.~Liu, W.~Yang, S.~Yang, and Z.~Guo, ``D3r-net: Dynamic routing residue
  recurrent network for video rain removal,'' \emph{IEEE Transactions on Image
  Processing (TIP)}, 2018.

\bibitem{yan2021self}
W.~Yan, R.~T. Tan, W.~Yang, and D.~Dai, ``Self-aligned video deraining with
  transmission-depth consistency,'' in \emph{Proceedings of the IEEE/CVF
  Conference on Computer Vision and Pattern Recognition}, 2021, pp.
  11\,966--11\,976.

\bibitem{yue2021semi}
Z.~Yue, J.~Xie, Q.~Zhao, and D.~Meng, ``Semi-supervised video deraining with
  dynamical rain generator,'' in \emph{Proceedings of the IEEE/CVF Conference
  on Computer Vision and Pattern Recognition}, 2021, pp. 642--652.

\bibitem{yang2020self}
W.~Yang, R.~T. Tan, S.~Wang, and J.~Liu, ``Self-learning video rain streak
  removal: When cyclic consistency meets temporal correspondence,'' in
  \emph{Proceedings of the IEEE/CVF Conference on Computer Vision and Pattern
  Recognition}, 2020, pp. 1720--1729.

\bibitem{jiang2017predicting}
L.~Jiang, M.~Xu, and Z.~Wang, ``Predicting video saliency with object-to-motion
  cnn and two-layer convolutional lstm,'' \emph{arXiv preprint
  arXiv:1709.06316}, 2017.

\bibitem{graves2005bidirectional}
A.~Graves, S.~Fern{\'a}ndez, and J.~Schmidhuber, ``Bidirectional lstm networks
  for improved phoneme classification and recognition,'' in \emph{International
  conference on artificial neural networks}.\hskip 1em plus 0.5em minus
  0.4em\relax Springer, 2005, pp. 799--804.

\bibitem{zhang2018residual}
Y.~Zhang, Y.~Tian, Y.~Kong, B.~Zhong, and Y.~Fu, ``Residual dense network for
  image super-resolution,'' in \emph{Proceedings of the IEEE conference on
  computer vision and pattern recognition}, 2018, pp. 2472--2481.

\bibitem{huang2017densely}
G.~Huang, Z.~Liu, L.~Van Der~Maaten, and K.~Q. Weinberger, ``Densely connected
  convolutional networks,'' in \emph{Proceedings of the IEEE conference on
  computer vision and pattern recognition}, 2017, pp. 4700--4708.

\bibitem{starik2003simulation}
S.~Starik and M.~Werman, ``Simulation of rain in videos,'' in \emph{Texture
  Workshop, ICCV}, vol.~2, 2003, pp. 406--409.

\end{thebibliography}

\end{document}